\documentclass[letterpaper, 10 pt, conference]{ieeeconf}  

\IEEEoverridecommandlockouts                              
\usepackage{amsmath,amsfonts}
\usepackage{algorithmic}
\usepackage{array}
\usepackage[caption=false,font=normalsize,labelfont=sf,textfont=sf]{subfig}
\usepackage{textcomp}
\usepackage{stfloats}
\usepackage{url}
\usepackage{verbatim}
\def\BibTeX{{\rm B\kern-.05em{\sc i\kern-.025em b}\kern-.08em
    T\kern-.1667em\lower.7ex\hbox{E}\kern-.125emX}}
\usepackage{balance}

\usepackage{graphicx}
\usepackage{epsfig}

\usepackage{footmisc}
\usepackage{etoolbox}

\title{\LARGE \bf
DiPA: Probabilistic Multi-Modal Interactive Prediction for Autonomous Driving

}

\author{Anthony Knittel$^{1}$, Majd Hawasly$^{1}$, Stefano V. Albrecht$^{1,2}$, John Redford$^{1}$, Subramanian Ramamoorthy$^{1,2}$
	\thanks{$^{1}$Five\,AI Ltd, UK. {\tt\small anthony.knittel@five.ai}}
	\thanks{$^{2}$School of Informatics, University of Edinburgh, Edinburgh, UK}%
}

\begin{document}

\maketitle

\begin{abstract}
Accurate prediction is important for operating an autonomous vehicle in interactive scenarios.
Prediction must be fast, to support multiple requests from a planner exploring a range of possible futures.
The generated predictions must accurately represent the probabilities of predicted trajectories, while also capturing different modes of behaviour (such as turning left vs continuing straight at a junction).
To this end, we present DiPA, an interactive predictor that addresses these challenging requirements.
%
Previous interactive prediction methods use an encoding of k-mode-samples, which under-represents the full distribution.
Other methods 
optimise closest-mode evaluations, which test whether one of the predictions is similar to the ground-truth, but allow additional unlikely predictions to occur, over-representing unlikely predictions.
%
DiPA addresses these limitations by using a Gaussian-Mixture-Model to encode the full distribution, and optimising predictions using both probabilistic and closest-mode measures.
These objectives respectively optimise probabilistic accuracy and the ability to capture distinct behaviours, and there is a challenging trade-off between them.  We are able to solve both together using a novel training regime.
%
%
%
DiPA achieves new state-of-the-art performance on the INTERACTION and NGSIM datasets, and improves over the baseline (MFP) when both closest-mode and probabilistic evaluations are used. 
This demonstrates effective prediction for supporting a planner on interactive scenarios.
\end{abstract}


\section{Introduction}

Prediction of the future motion of surrounding road users is essential for the safe operation of an autonomous vehicle (AV).
Road scenarios such as intersections, merges and roundabouts require significant interaction between agents in the scene, where agent behaviour is influenced by the presence of nearby agents, as well as reactions to actions that other agents take. 
In order to support planning, a predictor needs to estimate the future states of the surrounding road users based on observations of their recent history, and to estimate the risk of conflict for possible ego actions.

A planning system used in interactive scenarios needs to consider different possible actions that other vehicles may take, and the futures that result from different actions.  In order to explore these futures, a supporting predictor needs to be computationally fast, and to provide accurate predictions that represent the expected distribution of future states of each agent.
Many combinations of actions may be possible, so an interactive predictor needs to be fast in order to allow different 
futures to be explored.

Existing predictors addressing this task have encoded predictions using a fixed number of mode samples, for example using 6 predicted trajectories encoded as center positions~\cite{janjovs2021starnet, scibior2021imagining, gilles2021gohome}. These are evaluated using minimum average- or final-displacement error (minADE/FDE) and miss-rate (MR) (see Section~\ref{app:eval_details}). These measures compare the closest predicted mode with the ground-truth, and are important for demonstrating that predictions closely capture distinct modes of behaviour observed in the data.

\begin{figure}[t]
	\centering
	\includegraphics[width=\columnwidth]{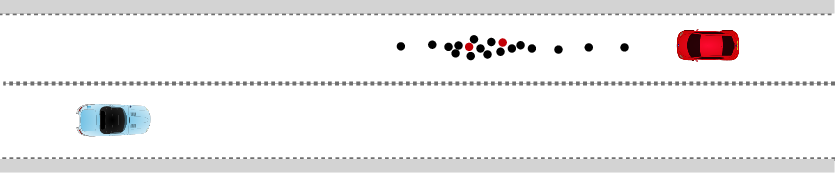}
	\includegraphics[width=\columnwidth]{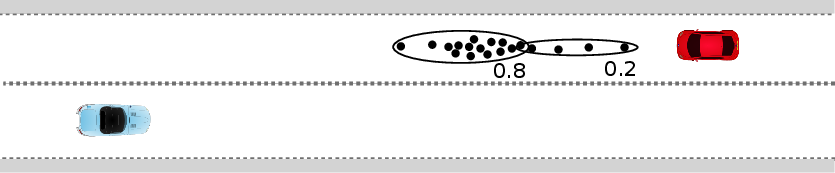}
	\caption{
    Top: 
    Use of k-mode samples (red, k=2) under-represents the distribution of future positions (black).  This prevents effective planning by underestimating states which are reasonably likely to occur. 
    Bottom: A GMM encoding, with associated mode weights, provides a more accurate representation of the full distribution by covering a wider range of samples.
	} 
	\label{fig_conflict_modes}
\end{figure}

\begin{figure}[t]
	\centering
	\includegraphics[width=\columnwidth]{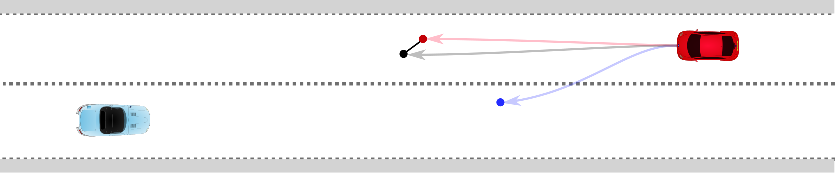}
	\includegraphics[width=\columnwidth]{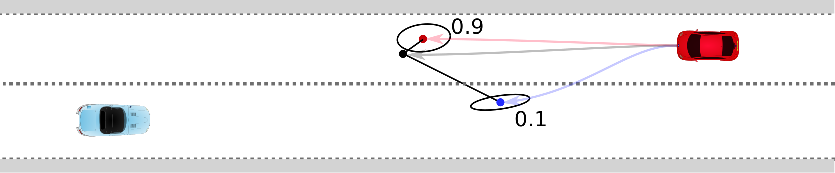}
	\caption{
    Top: Optimising for closest-mode evaluations can allow unrealistic predictions to be over-represented.
    For an instance of data 
    (black dot), the closest predicted mode (red) is evaluated while additional modes (blue) can predict unrealistic behaviours without penalty. 
    Unlikely predicted modes interfere with planning, for example causing an 
    emergency break to avoid a predicted collision that is unlikely.
    Bottom: Optimising for both closest-mode 
    and probabilistic evaluations 
    penalises unlikely predictions, 
    while minimising over- and under-representation. 
	} 
	\label{fig_conflict_evaluations}
\end{figure}

A limitation of this sample-based encoding is that it does not represent the full distribution of expected future positions, and as such many variations are under-represented (Fig.~\ref{fig_conflict_modes}).
A further limitation is that probabilities of predicted modes are not considered.  When training a model based on closest-mode evaluations, additional predicted modes (other than the closest) do not affect scoring, which allows the predictor to predict behaviour modes that are unlikely to occur. 
Each predicted mode has equal weight, which results in over representation of unlikely predictions (Fig.~\ref{fig_conflict_evaluations}).

These limitations can be addressed using a Gaussian Mixture Model (GMM), which 
represents the full predicted distribution, along with probability estimates of each mode. 
This is preferred over increasing the number of samples, as GMMs provide a compact encoding of the distribution and 
a practical means of evaluating the probability distribution.
Previous methods~\cite{mercat2020multi, tang2019multiple}
have used GMMs 
on the NGSIM dataset, which are evaluated using negative-log-likelihood (NLL) evaluations.  Further methods have used mode probability estimates~\cite{deo2018convolutional, antonello2022flash} which are evaluated using predicted-mode RMS (predRMS) evaluations (see Section~\ref{app:eval_details}). 

\begin{figure}[t]
	\centering
	\includegraphics[width=\columnwidth]{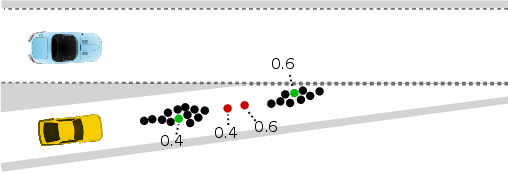}
	\caption{
	A merge scenario produces a bi-modal distribution (black samples). Optimising closest-mode (minADE/FDE) evaluations favours diverse predictions (green), while probabilistic (predRMS) evaluations favour predictions close to the mean (red), that minimise the penalty of incorrect mode estimates. 
	Solving both requires diverse predictions with the ability to 
	accurately estimate mode probabilities.
	} 
	\label{fig_eval_strategies}
\end{figure}

Probabilistic and closest-mode evaluations provide complimentary measures that are more informative than either alone, and are analogous to precision and recall in binary classification. 
We argue that an effective predictor for interactive scenarios needs to optimise both 
measures, to demonstrate that it is able to closely capture distinct behaviour modes, while also accurately representing probabilities. 
This is a challenging task as different evaluation measures are supported by contradictory prediction strategies.
Closest-mode evaluations (minADE/FDE/MR) favour diverse predictions,
while probabilistic evaluations (predRMS, NLL) favour conservative predictions close to the mean of expected behaviours, where the cost of 
incorrect mode estimates 
is minimised (Figure~\ref{fig_eval_strategies}).
Optimising 
both evaluation approaches together 
demonstrates accurate multi-modal prediction, and reduces the over-representation of unlikely predictions seen in 
Figure~\ref{fig_conflict_evaluations}.

To that end, we present DiPA (Diverse and Probabilistically Accurate) 
 -- a fast method for predicting in interactive scenarios using a GMM encoding, that is able to optimise both 
objectives 
together, by producing a diverse set of predictions with accurate probability estimates.  This allows distinct behaviours to be accurately modelled, while producing an accurate representation of the full trajectory distribution. 
This 
improves over previous methods~\cite{janjovs2021starnet, gilles2021gohome} using closest-mode evaluations on the INTERACTION dataset~\cite{zhan2019interaction}, and improves over previous methods~\cite{antonello2022flash, mercat2020multi} using probabilistic evaluations on NGSIM~\cite{ngsim}.  
DiPA also improves over a baseline method (Multiple-Futures Prediction (MFP))~\cite{tang2019multiple}
when comparing both closest-mode and probabilistic measures together. 
This demonstrates a predictor that is suitable for supporting an AV planner in interactive scenarios.

Beyond highlighting the importance of evaluating predictors with both closest-mode and probabilistic evaluations, 
the key contributions are: 
1) a fast prediction architecture with 
a flexible representation that processes agent interactions in wide-ranging road layouts, that produces high accuracy predictions on interactive scenarios,
2) a training regime that supports a diverse set of predicted modes using a GMM-based spatial distribution, with accurate probability estimates, and 
3) a revision to the NLL measure for evaluating GMM predictions, to correct for an important limitation.

\section{Related work}


A number of different structures have been used for prediction of agents in road scenes, including graph-, goal- and regression-based methods.  

StarNet~\cite{janjovs2021starnet} represents the 
scene and agents using vector-based graphs, and 
uses a combined representation of agents within their own reference frame 
and from the points of view of other agents.  Further graph-based methods such as \cite{mo2020recog, gilles2021gohome, jia2022multi} combine map information and agent positions into a common representation, commonly processed with a Graph Neural Network~\cite{zhou2020graph} 
in an encoder-decoder framework.  These methods allow encoding the static layout of the scene and various agents in a generalisable way, and have shown good results on closest-mode prediction.

Goal-based methods~\cite{zhao2020tnt, gu2021densetnt, hanna2021interpretable, albrecht2020igp2, brewitt2021grit} identify a number of potential future targets that each agent may head towards, determine likelihoods of each, and produce predicted trajectories towards those goals. 
Flash~\cite{antonello2022flash} uses a combination of Bayesian inverse-planning and mixture-density networks to produce accurate predictions of trajectories in highway driving scenarios.  
Goal-based methods 
use the map to inform trajectory generation, and can use kinematically-sound trajectory generators. However, this can lead to limited diversity on other factors such as motion profile and path variations compared to data-driven methods.

Regression-based methods use representations that directly map observations to predicted outputs.
SAMMP~\cite{mercat2020multi} produces joint predictions of the spatial distribution of vehicles, 
using a multi-head self-attention function to capture interactions between agents.  
Multiple-Futures Prediction (MFP)~\cite{tang2019multiple} models the joint futures of a number of interacting agents, 
using learnt latent variables for generating predicted future modes.  
Mersch et al.~\cite{mersch2021maneuver} present a temporal-convolution method for predicting interacting vehicles in a highway scenario where neighbouring agents are assigned specific roles based on relative positions to a central agent. 
These regression-based methods can be fast and accurate, but may have limited generalisability to different layouts when role-based representation of inputs is used.


Existing interactive prediction using the INTERACTION dataset have demonstrated good results based on closest-mode evaluations (minADE\,/\,FDE\,/\,MR) ~\cite{janjovs2021starnet, scibior2021imagining, gilles2021gohome}.  These have typically used a prediction encoding using a fixed number of modes, each represented as a trajectory sample.  Optimising closest-mode evaluations produces diverse predictions, which closely capture distinct modes of behaviour.  

Methods using the NGSIM dataset  have shown good results on probabilistic evaluations (predRMS, NLL)~\cite{mercat2020multi, mersch2021maneuver, tang2019multiple, antonello2022flash}.  These have used a range of encodings including k-mode-samples~\cite{antonello2022flash}, or GMM models~\cite{mercat2020multi, deo2018convolutional, tang2019multiple}.  Optimising these measures allows the probability distribution of predictions to be captured, however may not capture distinct behaviour modes closely.

The importance of balancing prediction diversity and probabilistic accuracy has been recognised in~\cite{mercat2020multi} which trains a GMM model with NLL loss and shows improved diversity against prior art measured by MR.  Rhinehart et al.~\cite{rhinehart2018r2p2} examined the generation of paths that are both diverse, to cover instances in the dataset, and precise, to minimise inconsistency with the data, using a specific cross-entropy term per objective. 
This balance is also addressed in generative CVAE models such as~\cite{yuan2021agentformer} which uses a trajectory sampler trained to extract diverse and plausible samples generated by the model.

To address the limitations of closest-mode and mean evaluation measures, \cite{rhinehart2018r2p2} propose the use of information-based cross-entropy evaluations.  The importance of evaluating with both 
displacement-error and NLL-based evaluations has also been recognised by~\cite{salzmann2020trajectron++}, for evaluating trajectories produced by a generative model. 
Measures of diversity and precision have also been explored by~\cite{casas2020implicit} using closest-mode and mean mode evaluations, performed on joint predictions of various agents in a scene.  A limitation of this approach is the lack of probabilistic weighting of predicted modes.

A useful interactive predictor requires (1) speed, which can be achieved by minimising unnecessary complexity; (2) an accurate encoding of the full probability distribution over trajectory predictions, as provided by a GMM; and (3) accurate predictions that capture distinct behaviour modes with an accurate distribution, as measured by closest-mode and probabilistic evaluations.
Addressing these factors together is challenging, as there are trade-offs between solving each, for example increasing diversity to capture distinct behaviours introduces a cost with estimating the probability distribution accurately.  
We demonstrate that the proposed DiPA method addresses this joint task, in a generalisable way that can be applied to the various scenes of interactive scenarios.  

\section{Proposed method}

\begin{figure*}
	\centering
	\includegraphics[width=0.9\textwidth]{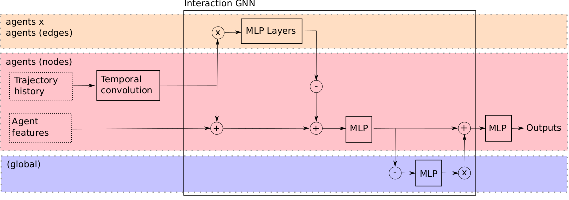}
	\caption{Network diagram of DiPA model. Trajectory history and agent dimensions are inputs.  The following symbols represent $\otimes$=broadcasting, $\oplus$=concatenation, $\ominus$=max reduction.  Outputs are predicted trajectories, spatial distribution parameters and mode prediction weights.}
	\label{fig_model_overview}
\end{figure*}

DiPA uses an encoder-decoder architecture, where interactions between agents are captured using a Graph Neural Network (GNN).  An overview of the network structure is shown in Figure~\ref{fig_model_overview}.   In order to support speed of processing, and to identify the essential elements needed, the design is focused on the minimal complexity that is needed to produce high-quality predictions.  
Agents are encoded based on observed histories such as positions and velocities,
and each of the agents are treated as symmetric entities in an unordered set, with no need to assign specific roles based on relative positions. 
This allows flexible comparisons between agents to be performed and enables generalisability to widely varying scenarios 
including roundabouts, junctions, highways and other road topologies with a varying numbers of agents in diverse arrangements.
Predictions are performed jointly on up to 20 agents at a time, while for evaluation purposes a single agent is used for each instance, where the surrounding neighbours are provided for context.

Inputs to the model are the observed histories of each agent (positions, orientations and speeds), and agent features including dimensions and type.  The model produces predictions as a multi-modal GMM, represented with a 2D Gaussian distribution for each timestep.  

Observed states for each agent are encoded using temporal convolution layers, 
and interactions between agents are processed using an edge-based GNN.  Edge features between pairs of interacting agents are produced by broadcasting agent encodings using concatenation, which are processed with MLP layers for each agent $\times$ agent pair.

This design has been chosen to emphasise the ability to directly process relative values between pairs of interacting agents, such as encodings of positions, velocities and orientations, which are trained based on regression. 
This is in contrast to standard GNN approaches~\cite{zhou2020graph}, which use an encoding per agent, where interactions between agents are processed using summation (or other reductions) of encoding messages passed from neighbouring nodes.
The proposed approach 
has similarities with the processing of entities in an unordered set used in PointNet~\cite{qi2017pointnet}. 


Reduction over edges (agent pairs) 
for each agent (node) 
produces a summary encoding for each agent, while reduction over agent nodes produces a scene context encoding, which allows properties of the scene to influence agent predictions.
%
%
The agent-context representation is decoded to produce predicted trajectory positions, spatial distribution parameters and mode weight estimates.  This design captures the important elements of processing agent predictions with interactions, while removing unnecessary complexity.  


\subsection{Training}

A typical approach for training GMM predictions is to minimise a NLL loss, such as the score used for evaluation in \eqref{eqn_nll}. When this loss is used, the spatial distribution parameters are updated using the predicted mode weight.  Inaccuracy in predicted weights produces randomness in mode training weights, 
resulting in mode convergence and loss of diversity.

We propose a novel training method that 
improves prediction diversity, which allows distinct modes of behaviour to be captured, and produces accurate estimates of probabilities.  
Training is performed with 1) a spatial distribution loss for training spatial distribution parameters, and 2) a mode weight estimation loss for training predicted mode weights.

Training mode weights 
(used with both losses) 
define the extent that each predicted mode will be updated based on an observation, where a flat training distribution leads to convergent modes while a biased distribution encourages diversity.  
Mode weight distributions $W\in\mathbb{R}^M, \sum^M_m W_m = 1$ 
represent the weighting of each mode for training or prediction.
Training mode weights $W_r$ are a combination of the closest mode weight $W_c$ and posterior mode weight $W_p$, using a proportion weighting $k_r=0.5$, chosen experimentally as described in Section~\ref{sec_ablation}.  
\begin{equation}
W_r = (1-k_r)W_c + k_rW_p \label{eqn_training_mode_weights}	
\end{equation}
$W_c$ is a strongly biased (one-hot) distribution that encourages training of the single most similar mode to the ground-truth. $\mu_{m,t}$ is the predicted trajectory position for mode $m$ at time $t$, and $x_t$ is the ground-truth.
\begin{align}
	W_{c,m} &=  \begin{cases}
		1 \quad \text{if} \, m = {\arg}{\min}_m (\frac{1}{T}\sum_t^T||x_t - \mu_{m,t}||) \\
		0 \quad \text{otherwise}
	\end{cases}
\end{align}

$W_p$ is a weakly biased distribution based on the posterior of the observation under the GMM model, and produces a balance of convergent and divergent mode training that facilitates participation of the different modes. 
$W_p$ prevents one or a few modes from dominating,
and reduces sensitivity to initialisation. $\Sigma_{m,t}$ is the predicted covariance matrix.
%
\begin{align}
	W_{p,m} &= \frac{1}{T}\sum_t^T \frac{\mathcal{N}(x, \mu_{m,t}, \Sigma_{m,t})}{\sum_i^M \mathcal{N}(x, \mu_{i,t}, \Sigma_{i,t})} \label{eqn_posterior}
\end{align}

In contrast, MFP~\cite{tang2019multiple} uses a combination of posterior and predicted distribution weights for training the 
GMM,  which has a tendency to produce a single dominant mode.  

\subsubsection{Spatial distribution training}

The spatial distribution loss \eqref{eqn_spatialloss} 
minimises the NLL score of an observation $x$ under the predicted model, weighted by the training mode distribution $W_r$.  This trains the parameters of the normal distribution $\mu, \Sigma$, while $W_r$ is constant. 
%
\begin{align}
	\mathcal{L}_{spatial} &= -\frac{1}{T} \sum_t^T \ln(\sum_m^M W_{r, m} \mathcal{N}(x, \mu_{m,t}, \Sigma_{m,t})) \label{eqn_spatialloss}
\end{align}

The training weight distribution $W_r$ 
emphasises training modes similar to the observation, supporting mode diversity, in contrast to standard NLL training based on the predicted mode weight distribution.

\subsubsection{Mode weight estimation training}
Two predicted mode weight terms 
are produced by the model, $W_s$ and $W_n$, which are based on similarity of trajectory positions, 
and low spatial distribution error respectively.  
Separate terms are used as the ideal mode weights can be inconsistent for different objectives.
The trajectory-based mode estimation weight $W_s$ is trained with a MSE-based loss, 
as shown in \eqref{eqn_moderms}.  
($W_s$ is trained while $\mu$ is constant)
%
\begin{equation}
    \mathcal{L}_{MSE} = \sum_m^M W_{s,m} \frac{1}{T}\sum_t^T||x_t - \mu_{m,t}||^2\label{eqn_moderms}	
\end{equation}

The spatial distribution mode weight $W_n$ is trained in order to 
minimise the NLL score, and 
to approach the training mode distribution $W_r$, as shown in~\eqref{eqn_modedist}.  
($W_n$ is trained and $\mu, \Sigma, W_r$ are constant)
%
\begin{multline}
	\mathcal{L}_{DIST} = -\frac{1}{T} \sum_t^T \ln(\sum_m^M W_{n, m} \mathcal{N}(x, \mu_{m,t}, \Sigma_{m,t}))\\
	+ D_{KL}(W_r || W_n) \label{eqn_modedist}	    
\end{multline}

%

The two mode estimation distributions $W_s$ and $W_n$ are based on different objectives, and favour
trajectory- and distribution-based evaluations respectively.  
In order to produce a single prediction that balances these objectives, 
a weighted average is returned $W_o = (1-k_n)W_s + k_nW_n$, using $k_n=0.9$, which has been chosen experimentally as shown in Section~\ref{sec_ablation}, so that the proposed method out-performs prior methods on all tasks.

\section{Experiments}
Experiments are conducted to demonstrate that the proposed DiPA method meets the objectives for supporting an interactive planner, which requires fast processing, the ability to capture distinct modes of behaviour, and to accurately capture the probability distribution for predictions.  
Experiments are conducted on the INTERACTION~\cite{zhan2019interaction} dataset to compare 
existing benchmarks using closest-mode evaluations, to demonstrate capturing distinct behaviour modes. Experiments on NGSIM~\cite{ngsim} compare against prior methods using probabilistic evaluations, to demonstrate the ability to capture the distribution accurately.  As there is a trade-off between optimising closest-mode and probabilistic tasks, experiments are conducted on the joint task using both evaluation approaches on each dataset, which is compared against MFP as a baseline (Section~\ref{app:mfp_revised}).



\subsubsection{INTERACTION dataset} The INTERACTION dataset~\cite{zhan2019interaction} is divided into instances based on each fully-observed agent in each case window, with a 4 second duration. 
Prediction is performed using a 1 second observed period and 3 second prediction period.


\subsubsection{NGSIM dataset} The NGSIM dataset contains trajectory tracks for agents in two scenes (US-101 and I-80). Agents are assigned to train/evaluation splits based on vehicle identifier, as used in~\cite{deo2018convolutional}.  Instances are created based on a central agent, for each fully-observed window of 8 seconds (3 observed, 5 future).  For each instance up to 20 neighbouring agents are also observed, while agents that have been assigned to different splits are not used for training.

On both datasets global coordinates are used. 
Pre-processing centers units on the last observed position of the agent to be predicted, 
with rotation such that the yaw of the prediction agent is zero (at the last observed timestep).  

%

\subsection{Revised implementation of Multiple-Futures Prediction}
\label{app:mfp_revised}


MFP~\cite{tang2019multiple} is a useful baseline as it is an accurate method based on a GMM, allowing comparison on each of the evaluation measures.  
A limitation of MFP is that it has been implemented using local lane-based coordinates, which are suitable for highway driving involving mostly parallel lanes.  This representation is not directly generalisable to more complex scenarios involving intersections, roundabouts and other non-parallel topology, as are used in INTERACTION.

In order to use global coordinates, for consistency each instance is re-framed to be centered on the last observed position and rotated on the orientation of the central agent.  
A revised neighbour grid is used to allow MFP to operate on 
widely varying road topologies.
MFP represents neighbours using a $13\times3$ grid of positions in the central and neighbouring lanes, based on distances from the central agent.  
A comparable neighbour grid is produced based on the central and neighbouring lane patches corresponding with the central agent.  All following and preceding lane patches from the central lane patch(es) represent the central lane, and similarly for the neighbouring lanes.
Grid spacing distances for each neighbour agent are found based on the nearest midline path, using the progress distance of the neighbour agent relative to the central agent. 
This defines a neighbour grid similar to that used in MFP, and implements the \textit{MFP-general} method. 

\subsection{Evaluation measures}
\label{app:eval_details}


\subsubsection{predRMS} the RMS error of the most probable predicted mode is calculated for a number of timesteps, over the instances of the dataset $N$ as shown in \eqref{eqn_predrms}, where $\mu_i$ is the predicted position for the most probable mode $i=\underset{m}{\arg \max}(W_m)$
as used in~\cite{deo2018convolutional, tang2019multiple, antonello2022flash}: 
\begin{align}
	predRMS_t = \sqrt{\frac{1}{N}\sum_{n=1}^N||x_{n,t}-\mu_{i,t}||^2}\; \label{eqn_predrms}
\end{align}
We use the same number of modes as used in corresponding minADE/FDE experiments.

\subsubsection{minADE} evaluates the closest average Euclidian distance between the predicted trajectory mode and the ground truth over a horizon $T$, while \textit{minFDE} evaluates the closest final position, 
as follows.
\begin{align}
	minADE &= \min_m(\frac{1}{T}\sum_{t=1}^T||x_t-\mu_{t,m}||)\\
	minFDE &= \min_m(||x_T-\mu_{T,m}||) \label{eqn_minade}
\end{align}

\subsubsection{Miss-rate (MR)} is defined as the percentage of instances where the minimum spatial error on the final timestep is larger than a given threshold, ie $minFDE > k\in\mathbb{R}$.  We use a threshold of $k=2m$ as used in~\cite{zhao2020tnt, mercat2020multi}. 

\subsubsection{Negative-log-likelihood (NLL)} describes the log-probability of observed instances under a predicted distribution.
Previous methods~\cite{tang2019multiple, song2020pip, mercat2020multi} use a GMM representation, although NLL 
can be compared between different representations.
Calculation of the NLL score using a GMM is shown in \eqref{eqn_nll}. This is represented using a center position $\mu_m\in\mathbb{R}^2$, covariance matrix $\Sigma_m\in\mathbb{R}^{2\times2}$ and weight $W_m\in\mathbb{R}$ for each predicted mode, where $x\in\mathbb{R}^2$ is the ground-truth position.  

\begin{align}
	NLL &= \frac{1}{T}\sum_{t=1}^T -\ln(\sum_{m=1}^MW_m\mathcal{N}(x_t, \mu_{m,t}, \Sigma_{m,t})) \label{eqn_nll}\nonumber\\
	\mathcal{N}(x, \mu, \Sigma) &= \frac{1}{2\pi\sqrt{|\Sigma|}}e^{-\frac{1}{2} (x-\mu)^T\Sigma^{-1}(x-\mu)}
\end{align}


\label{app:nll_details}

NLL as a concept is a dimensionless property, however previous results and the evaluation in \eqref{eqn_nll} represent 
probability density, 
without reducing to a dimensionless value. 
Observed samples are points, which have zero probability in a spatial distribution as a result of being a position with no size.  It is possible to produce a probability evaluation using an area instead of a point, 
however the area to use is not well defined or meaningful for the task. 
As dimensioned probability density values are used, and the NLL measure is determined from this value, it is important to record the units of the density-based NLL property reported.
Previous methods have used inconsistent units, 
in feet~\cite{tang2019multiple} and in meters~\cite{mercat2020multi}, 
so to address this problem we present units of measurement with reported results. 

Another limitation is that in existing definitions NLL is an unbounded quantity, which allows scores on a small number of instances to greatly influence evaluation over the dataset.  
This is both a theoretical and a practical problem, as for example a dataset may contain a stationary object, where the center of a predicted GMM can be accurately chosen, and the distribution width reduced to an arbitrarily high density,
bounded only by numerical limits.  
When represented with a 64-bit float (with limit $5.5 \times 10^{-309}$), this can result in a NLL score of $-710$ for a single instance.

We suggest that a maximum probability density be applied, as for vehicle prediction there is no practical advantage in distinguishing between very tight bounds.  Mercat et al.~\cite{mercat2020multi} apply a minimum limit to the standard deviation 
of $\sigma=0.1m$, for the purposes of avoiding overfitting. 
We 
extend this definition to apply to evaluation, where the probability density is capped for each instance, based on the maximum probability density of a normal distribution with $\sigma=0.1m$, 
which gives a minimum NLL score of $-\ln(\frac{1}{2\pi0.1^2})$ (approx. $-2.77$).  
This can be used with any probability distribution, including GMMs and raster-based representations.

\section{Results}

\paragraph{Capturing probability distribution} Comparison using probabilistic evaluations on NGSIM are shown in Tables~\ref{table_ngsim_predrms} and \ref{table_ngsim_nll}
(ablations are below the double line, as discussed in Section~\ref{sec_ablation}).
DiPA improves over previous methods on predRMS evaluations, which involves generating a set of predicted trajectories and accurately predicting the most probable mode.  
Evaluation of the spatial distribution using NLL 
shows improved probabilistic accuracy with DiPA over previous methods.
These experiments show advantages of DiPA 
for capturing probabilistic predictions on NGSIM.

\begin{table}[t]
	\centering
	\small
	\caption{\small Probabilistic predRMS scores on NGSIM}
 	\begin{minipage}{\columnwidth} 
    	\makebox[\linewidth]{
    		\begin{tabular}{|l|r r r r r|}
    			\hline
    			& \multicolumn{5}{c|}{predRMS (by time period) $[m]$}\\
    			Method                                  & 1s   & 2s   & 3s   & 4s   & 5s \\
    			\hline
    			CV~\cite{mercat2019kinematic}         & 0.76    & 1.82    & 3.17    & 4.80    & 6.70  \\
    			CSP(M)~\cite{deo2018convolutional, mercat2020multi}      & 0.59    & 1.27    & 2.13    & 3.22    & 4.64 \\
    			GRIP\footnote{Results are adjusted to correct for scoring with RMS with average over spatial dimension values instead of Euclidean distance RMS.}~\cite{li2019grip}      & 0.52 & 1.22 & 2.05 & 3.13 & 4.47 \\ 
    			SAMMP~\cite{mercat2020multi}           & 0.51 & 1.13 & 1.88 & 2.81 & 3.98 \\
    			Flash~\cite{antonello2022flash}        & 0.51 & 1.15 & 1.84 & 2.64 & 3.62 \\
    			MFP~\cite{tang2019multiple}
                & 0.54 & 1.17 & 1.87 & 2.71 & 3.67 \\
    			\textbf{DiPA}                          & \textbf{0.46}  & \textbf{1.04}  & \textbf{1.70}  & \textbf{2.47}   & \textbf{3.43}   \\
    			\hline
    			\hline
    			Trajectory mode weight    & 0.46  & 1.04 & 1.70 & 2.45 & 3.39 \\
    			Spatial mode weight    & 0.47  & 1.08 & 1.79 & 2.62 & 3.64 \\
    			Standard NLL loss & 0.44 & 1.03  & 1.66 & 2.48 & 3.50 \\
    			Closest-mode training & 0.43 & 0.99    & 1.64 & 2.44 & 3.43\\
    			Posterior training & 0.43 & 0.99  & 1.65 & 2.47 & 3.50 \\
    			\hline
    		\end{tabular}
    	}
    \end{minipage}
	\label{table_ngsim_predrms}
\end{table}

\begin{table}[t]
	\centering
	\small
	\caption{\small Probabilistic NLL scores on NGSIM (modes=5)}
  	\begin{minipage}{\columnwidth} 
    	\makebox[\linewidth]{
    		\begin{tabular}{|l|r r r r r|}
    			\hline
    			& \multicolumn{5}{c|}{NLL (by time period) $[\ln m^{-2}]$}\\
    			Method                                  & 1s   & 2s   & 3s   & 4s   & 5s  \\
    			\hline
    			CV~\cite{mercat2019kinematic}\footnote{
                Previously reported NLL results do not use the thresholded NLL score described in Section~\ref{app:nll_details}. 
                \label{nllunthresholded}
                }         & 0.82 & 2.32 & 3.23 & 3.91 & 4.46 \\
    			CSP(M)~\cite{deo2018convolutional, mercat2020multi}\footref{nllunthresholded}      & -0.41 & 1.07 & 1.93 & 2.55 & 3.08 \\
    			SAMMP~\cite{mercat2020multi}\footref{nllunthresholded}           & -0.36 & 0.70 & 1.51 & 2.13 & 2.64 \\
    			MFP~\cite{tang2019multiple}
                & -0.64    & 0.71  & 1.56  & 2.21  & 2.74 \\
    			\textbf{DiPA}                          &\textbf{-1.22}  & \textbf{0.20}  & \textbf{1.23}  & \textbf{2.01}  & \textbf{2.61} \\			
    			\hline
    			\hline
    			Non-thresholded    & -2.50 & -0.14 & 1.12 & 1.98 & 2.60 \\
    			Trajectory weight    & 9810.85 & 3670.04 & 63.28 & 29.92 & 17.52 \\
    			Spatial weight    & -1.24 & 0.18 & 1.21 & 2.00 & 2.60 \\
    			Standard NLL loss & -1.36 & 0.06  & 1.11 & 1.91 & 2.60\\
    			Closest-mode  & -1.17 & 0.51    & 1.60 & 2.36 & 2.86\\
    			Posterior training & -1.30 & 0.11  & 1.16 & 1.95 & 2.60 \\
    			\hline
    		\end{tabular}
    	}
    \end{minipage}
	\label{table_ngsim_nll}
\end{table}

\paragraph{Capturing distinct behaviours} Comparisons 
using closest-mode evaluations on INTERACTION is shown in
Table~\ref{table_interaction_closest}. 
DiPA shows lower error based on the closest mode than the comparison methods. 
Comparison of closest-mode Miss-Rate (MR) evaluations on NGSIM are shown in Table~\ref{table_ngsim_closest}.
DiPA shows improved MR evaluation over methods such as SAMMP~\cite{mercat2020multi} that are based on standard NLL training, 
showing improved ability to produce diverse modes that closely cover individual instances of the dataset.
These experiments show that DiPA improves over previous methods for accurately capturing distinct modes of behaviour.

\begin{table}[t]
	\caption{\small Closest-mode scores on INTERACTION (modes=6)}
	\begin{minipage}{\columnwidth} 
		\centering
		\small
		\begin{tabular}{|l|r r r|}
			\hline
			Method & minADE$_6$ & minFDE$_6$ & MR$_6$ \\
			\hline
			TNT~\cite{zhao2020tnt}               & 0.21   & 0.67 & \_ \\
			ReCoG~\cite{mo2020recog}             & 0.19   & 0.66 & \_ \\
			ITRA~\cite{scibior2021imagining}     & 0.17   & 0.49 & \_ \\
			GoHome~\cite{gilles2021gohome}       & \_       & 0.45 & \_ \\
			StarNet~\cite{janjovs2021starnet}
			& 0.16   & 0.49 & \_ \\
			joint-StarNet~\cite{janjovs2021starnet} 
			& 0.13   & 0.38 & \_ \\
			MFP-general              & 0.43      & 1.20    & 0.19\\
			\textbf{DiPA} & \textbf{0.11}    & \textbf{0.34}    & \textbf{0.02}\\
			\hline
			\hline
			Trajectory mode weight & 0.11 & 0.34 & 0.02\\
			Spatial mode weight & 0.11 & 0.34 & 0.02 \\
			Standard NLL loss &  0.18    & 0.47    & 0.03\\
			Closest-mode training & 0.11 & 0.33    & 0.01\\
			Posterior training & 0.11 & 0.36    & 0.02\\		
			\hline
		\end{tabular}
	\end{minipage}
	
	\label{table_interaction_closest}
\end{table}

\begin{table}[t]
	\centering
	\small
	\caption{\small Closest-mode scores on NGSIM}
	\makebox[\linewidth]{
		\begin{tabular}{|l|r r r|}
			\hline
			Method & minADE$_5$ & minFDE$_5$ & MR$_5$\\
			\hline
			CV~\cite{mercat2019kinematic}         & \_& \_ & 0.71\\
			CSP(M)~\cite{deo2018convolutional, mercat2020multi}      &  \_& \_ & 0.44\\
			SAMMP~\cite{mercat2020multi}           & \_& \_ & 0.23\\
			MFP~\cite{tang2019multiple}                            & 1.07   & 2.15   & 0.40  \\
			\textbf{DiPA}                          &\textbf{0.48}   & \textbf{0.86}   & \textbf{0.07}  \\			
			\hline
			\hline
			Trajectory mode weight    &  0.48&0.86 &0.07 \\
			Spatial mode weight    & 0.48& 0.86& 0.07\\
			Standard NLL loss & 0.90 & 1.75  & 0.32 \\
			Closest-mode training & 0.46 & 0.82    & 0.05 \\
			Posterior training & 0.51 & 0.99  & 0.16 \\
			\hline
		\end{tabular}
	}
	\label{table_ngsim_closest}
\end{table}

\paragraph{Combined task} In order to compare the ability to optimise both closest-mode and probabilistic tasks at the same time, results using multiple evaluation measures 
are shown for MFP and DiPA in
Tables~\ref{table_ngsim_predrms},~\ref{table_ngsim_nll} and \ref{table_ngsim_closest}. These show that MFP produces accurate probabilistic predictions as measured with predRMS and NLL, however shows relatively high error on closest-mode evaluations.  This suggests limited diversity of predictions, which limits the ability to closely match individual instances.  
Comparison of multiple evaluations on INTERACTION is shown in Tables~\ref{table_interaction_closest},~\ref{table_interaction_predrms} and \ref{table_interaction_nll}. This also shows improved results with DiPA against MFP-general on all evaluation measures,
showing advantages of DiPA for optimising both tasks together. 


\begin{table}[t]
	\centering
	\small
	\caption{\small Probabilistic predRMS scores on INTERACTION}
	\begin{tabular}{|l|r r r|}
		\hline
		& \multicolumn{3}{c|}{predRMS $[m]$}\\
		Method & 1s & 2s & 3s \\
		\hline
		MFP-general              & 0.21  & 0.95  & 2.37\\
		\textbf{DiPA} & \textbf{0.11}  & \textbf{0.47}  & \textbf{1.28}\\
		\hline
		\hline
		Trajectory mode weight & 0.11  & 0.47  & 1.25\\
		Spatial mode weight & 0.12  & 0.51  & 1.38\\
		Standard NLL loss& 0.11  & 0.44  & 1.18  \\
		Closest-mode training & 0.15  & 0.50  & 1.28  \\
		Posterior training & 0.10  & 0.46  & 1.27  \\
		\hline
	\end{tabular}
	\label{table_interaction_predrms}
\end{table}

\begin{table}[t]
	\centering
	\small
	\caption{\small Probabilistic NLL scores on INTERACTION (modes=6)}
	\begin{tabular}{|l|r r r|}
		\hline
		& \multicolumn{3}{c|}{NLL $[\ln m^{-2}]$}\\
		Method & 1s & 2s & 3s \\
		\hline
		MFP-general              & -1.87  & 0.46  & 2.17 \\
		\textbf{DiPA} & \textbf{-2.09}  & \textbf{-0.85}  & \textbf{0.76}\\
		\hline
		\hline
		Non-thresholded & -4.82  & -1.67  & 0.35 \\
		Trajectory mode weight & 93.20  & 18.19  & 12.79 \\
		Spatial mode weight & -2.10  & -0.87  & 0.76 \\
		Standard NLL loss & -2.27  & -0.58  & 0.95 \\
		Closest-mode training & -1.56  & -0.87  & 0.73 \\
		Posterior training & -2.22  & -0.94  & 0.70 \\
		\hline
	\end{tabular}	
	\label{table_interaction_nll}
\end{table}

\begin{figure*}
	\centering
	\includegraphics[scale=.2]{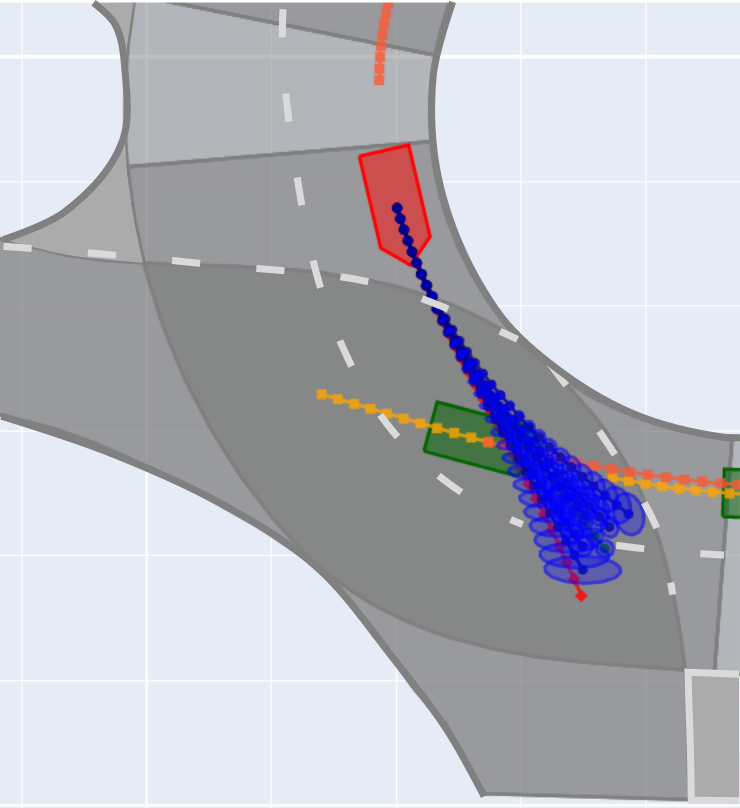}\hfill
	\includegraphics[scale=.2]{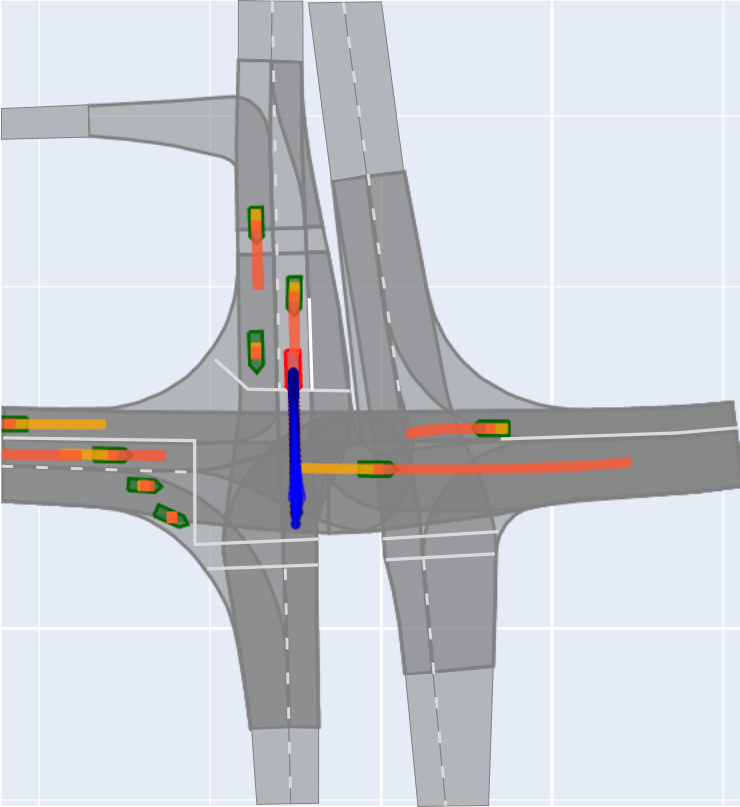}\hfill
	\includegraphics[scale=.2]{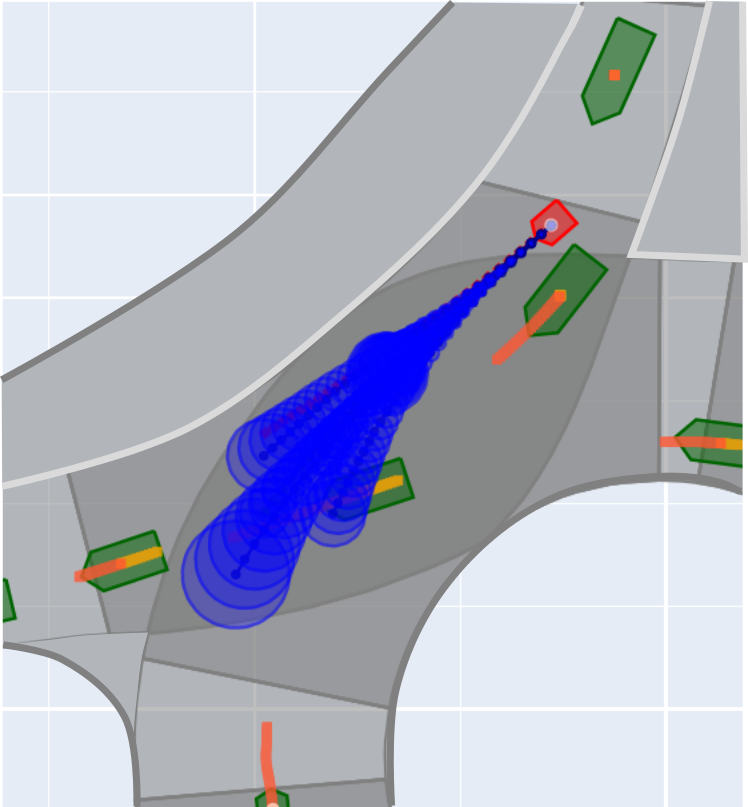}\hfill
	\includegraphics[scale=.2]{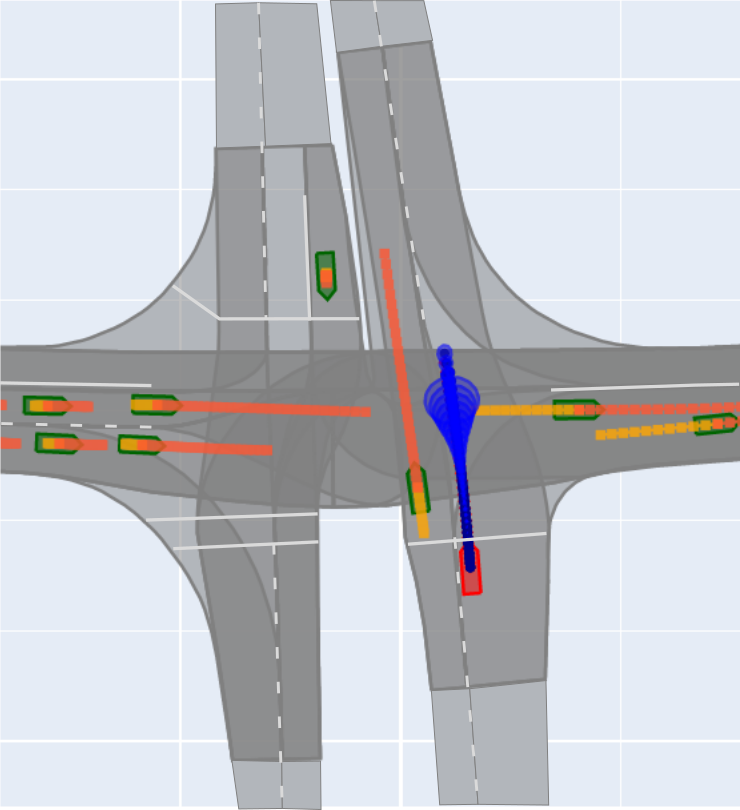}
	\caption{
	Qualitative results, showing use of the GMM encoding to represent the full predicted distribution (blue ellipses show $\sigma=1$). Ego is red vehicle, green vehicles are neighbours, showing history (yellow) and future (orange). L-R: 1. spread to capture variations over chosen paths, 2. narrow spread with variations in speed when crossing intersection, 3. cyclist prediction with large variations in speed and path, 4. distinct modes, with narrow prediction while crossing intersection and also wide spread at slower speeds.  
	 }
	\label{fig_qualitative}
\end{figure*}

Figure~\ref{fig_qualitative} shows selected instances from the INTERACTION dataset and predictions produced by DiPA.  This shows DiPA's ability to represent the full predicted distribution using a GMM encoding, 
providing greater coverage of variations compared to the typical k-mode-samples approach, 
while also capturing distinct modes of behaviour. 

The run-time of the model is 16ms per call 
to predict 20 agents at time, using a NVidia 2080Ti GPU.  This fast run-time for repeat calls allows multiple predictions to be made as part of inference performed by a planner.

\subsection{Ablation study}
\label{sec_ablation}

\begin{figure}
	\centering
	\includegraphics[width=.37\textwidth]{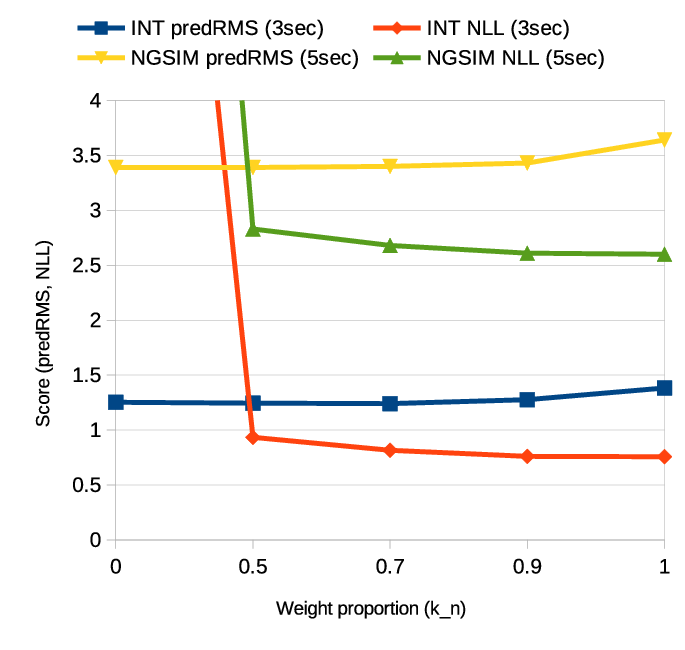}
	\caption{
	Effect of changing proportion $k_n$ balancing trajectory- and spatial-distribution-based mode weights.}
	\label{fig:proportion}
\end{figure}

Experiments using variations of DiPA are shown in each result table below the double line.  Evaluating with \textit{Non-thresholded} NLL scores show lower error values, particularly for short time horizons that involve narrower error distributions. These low scores can result from tight bounds on a few instances, and thresholded scores are more informative. 
Predicting with the \textit{trajectory mode weight} $W_s$ alone ($k_n=0$) favours RMS scores at a cost of NLL evaluations, while the \textit{spatial mode weight} $W_n$ alone ($k_n=1$) produces lower NLL error with increased RMS errors.
The effect of changing the proportion $k_n$ 
is shown in Figure~\ref{fig:proportion}, showing values for the final timestep.  A proportion of $k_n=0.9$ 
allows effective prediction according to both trajectory- and distribution-based evaluation.
\textit{Standard NLL loss} shows a condition where training is performed directly from the NLL loss, using the predicted mode distribution as the training weights. 
This shows lower error on NLL, but substantially higher error on closest-mode evaluations, showing that it is not able to capture specific behaviour modes as well.  
\textit{Closest-mode training} based on the mode weight $W_c$ only ($k_r=0$) shows lower closest-mode errors but higher NLL error on NGSIM, which contains more noise than INTERACTION.  During development, we have found that using $W_c$ only can lead to one or a few modes dominating, and is expected to be sensitive to initialisation. \textit{Posterior training} based on $W_p$ ($k_r=1$) shows lower predRMS and NLL error but higher closest-mode error, showing it is not as effective at capturing distinct behaviour modes.  
A balanced setting ($k_r=0.5$) 
provides a reliable approach that improves over prior methods on all measures.



\section{Conclusion}

In order to support an AV planner for operating in interactive scenarios, a predictor needs to be fast, to identify distinct modes of behaviour, and accurately represent the probability distribution of predictions.  Previous interactive predictors are able to capture distinct modes of behaviour, as measured by closest-mode evaluations, however the k-mode-samples encoding under-represents the full distribution, and misses many variations that can reasonably be expected.  In addition, 
when probability estimates are not reported or used for optimisation, the predictions can over-represent behaviours that are unlikely to occur.

Our proposed DiPA method uses a GMM encoding to represent the full predicted distribution, and uses a novel architecture and training regime that allows learning of distinct modes of behaviour, while also accurately representing the probability distribution.  Solving both of these tasks together is more challenging than solving either on its own.



Results on the INTERACTION and NGSIM datasets show DiPA captures distinct behaviour modes and probability estimates better than previous methods. 
There is a trade-off between these tasks, and comparison using both 
evaluations shows improvement over the MFP baseline on each measure, demonstrating the ability to accurately model the probability distribution of a diverse set of predicted behaviours. 


Limitations of DiPA are that it does not use map information, which prevents following a given road layout, 
and as a regression-based network, the model can also occasionally produce unrealistic predictions. This is currently mitigated using a wrapper to constrain maximum predicted speeds.

DiPA shows fast run times, and produces an accurate encoding of the full distribution of predictions, that minimises both over- and under-representation of predictions.  This provides useful predictions for supporting an AV planner in interactive scenarios.



\bibliography{intpred}  
\bibliographystyle{IEEEtran}

\end{document}